\documentclass[graybox]{svmult}

\usepackage{mathptmx}       
\usepackage{helvet}         
\usepackage{courier}        
\usepackage{type1cm}        
\usepackage{makeidx}         
\usepackage{graphicx}        
\usepackage{multicol}        
\usepackage[bottom]{footmisc}
\usepackage{algorithm}
\usepackage{algorithmic}
\usepackage{inconsolata}
\usepackage{times}
\usepackage{latexsym}
\usepackage{svg}
\usepackage{tabularx}
\usepackage[T1]{fontenc}
\usepackage[utf8]{inputenc}
\usepackage{microtype}
\usepackage{mathtools}
\usepackage{multirow}
\usepackage{arydshln}
\usepackage{newfloat}
\usepackage{listings}
\usepackage{newtxtext}  
\usepackage{newtxmath}  

\makeindex             

\begin{document}

\title*{Enhancing Robustness of LLM-Synthetic Text Detectors for Academic Writing: A Comprehensive Analysis}
\titlerunning{Enhancing Robustness of LLM-Synthetic Text Detectors for Academic Writing}
\author{Zhicheng Dou$^{1,2*}$, Yuchen Guo$^{1,2*}$, \\
Ching-Chun Chang$^{1}$, Huy H. Nguyen$^{1}$, and Isao Echizen$^{1,2}$ \\
\small{$^{1}$National Institute of Informatics, Japan \ \ \ \ \ \ \ \ $^{2}$The University of Tokyo, Japan} \\
\small{$^{*}$These authors contributed equally} \\
\small{E-mail: \{dou, guoyuchen, ccchang, nhhuy, iechizen\}@nii.ac.jp}
}
\authorrunning{Zhicheng Dou and Yuchen Guo \textit{et al.}}
%
%
\maketitle
\abstract{The emergence of large language models (LLMs), such as Generative Pre-trained Transformer 4 (GPT-4) used by ChatGPT, has profoundly impacted the academic and broader community. While these models offer numerous advantages in terms of revolutionizing work and study methods, they have also garnered significant attention due to their potential negative consequences. One example is generating academic reports or papers with little to no human contribution. Consequently, researchers have focused on developing detectors to address the misuse of LLMs. However, most existing methods prioritize achieving higher accuracy on restricted datasets, neglecting the crucial aspect of generalizability. This limitation hinders their practical application in real-life scenarios where reliability is paramount. In this paper, we present a comprehensive analysis of the impact of prompts on the text generated by LLMs and highlight the potential lack of robustness in one of the current state-of-the-art GPT detectors. To mitigate these issues concerning the misuse of LLMs in academic writing, we propose a reference-based Siamese detector named \textbf{Synthetic-Siamese} which takes a pair of texts, one as the inquiry and the other as the reference. Our method effectively addresses the lack of robustness of previous detectors (OpenAI detector and DetectGPT) and significantly improves the baseline performances in realistic academic writing scenarios by approximately 67\% to 95\%.}

\section{Introduction}
Large-scale language models (LLMs), such as OpenAI's GPT-4~\cite{openaigpt4}, and Google's Pathways Language Model 2~\cite{anil2023palm}, have become an integral part of our lives and jobs and are often utilized unknowingly. However, while LLMs greatly facilitate daily activities, they also pose significant security risks if maliciously exploited for attacks or deception. Consequently, with the growing popularity of LLMs, the importance of AI security has come to the forefront of people's attention~\cite{crothers2023machine, disinformation, greshake2023youve}. Among the various security concerns, academic cheating stands out as a particularly grave issue. ChatGPT, in particular, has gained widespread popularity among college students worldwide. Consequently, universities urgently need robust detectors, which has driven continuous advancements in the field of detection technology.

Researchers in detector development have explored strategies to optimize the training set for improved model performance. Notably, Liyanage et al.~\cite{liyanage2022benchmark} pioneered an AI-generated academic dataset using GPT-2, although it is considered less effective than the more advanced ChatGPT model currently available. Yuan et al.~\cite{yuan2021bartscore} proposed BERTscore, an evaluation method for filtering high-quality generated text that closely resembles human writing. Such text can be incorporated into the training set, thereby enhancing the performance of the detectors.

Researchers have also focused on optimizing the detector itself. Jawahar et al.~\cite{jawahar2020automatic} addressed the challenge of hybrid text, introducing a method to detect the boundary between machine-generated and human-written content, rather than solely distinguishing between the two. Zhao et al.~\cite{zhao2023survey} conducted a comprehensive survey of various LLMs, analyzing their performance across multiple dimensions, including pretraining, adaptation tuning, utilization, and capacity evaluation. They also identified potential future development directions for LLMs. Additionally, Mitchell et al.~\cite{mitchell2023detectgpt} proposed a model utilizing a curvature-based criterion to determine whether a given passage was generated by an LLM.

Studies examining the robustness of detectors include Rodriguez et al.~\cite{fine-tunning}, which investigated the impact of dataset domain on detector performance, highlighting a significant decrease in performance when the training and test datasets differ in domain. Their findings emphasized how the diversity of training sets directly affects the detector's performance. In addition, Pu et al.~\cite{pu2022deepfake} analyzed the issue of insufficient robustness in existing detection systems by exploring changes in decoding or text sampling strategies. 
While previous research focused on robustness in terms of dataset domains and generation models' parameters, \textbf{this study highlights that prompt adjustments alone can significantly affect the robustness of the detector, particularly in the context of academic cheating}.

The contributions of our paper are as follows:

\begin{itemize} 
\item \textbf{Highlighting the insufficient robustness of existing detectors through the example of academic writing cheating:} We demonstrate that solely adjusting the prompts is inadequate for ensuring the robustness of existing detectors (OpenAI detector and DetectGPT), particularly in the context of academic writing cheating.

\item \textbf{Exploring the prompt-induced lack of robustness and evaluating model applicability:} We put forward a hypothesis to explain the reasons behind the lack of robustness caused by input prompts and provide a theoretical basis for our following experiments.

\item \textbf{Introducing Synthetic-Siamese, an approach for detecting cheating in academic writing:} We analyze cheating in academic writing and propose a detection approach based on a Siamese network. Synthetic-Siamese exhibits greater prompt generalization capabilities compared to existing detectors, effectively addressing the issue of insufficient robustness.
\end{itemize}


\section{Asserting the Limitation of Existing Detectors}
\label{sec:limitation}
Since the release of GPT-3, OpenAI has allowed users to provide input prompts to shape the output text, enabling a wide range of functionalities. 
In contrast, the previous model, GPT-2, lacks prompt functionality and is irrelevant to the robustness of prompt-related issues. ChatGPT, a question-answering platform, does not offer adjustable parameters. Therefore, the output hardly changes when using the same prompt, making it unsuitable for generating large-scale datasets with diverse outputs. GPT-4, the latest LLM at the time of writing, has shown to be highly effective in terms of text quality and multi-modality. However, since GPT-4 was still in development when conducting this experiment, we did not choose GPT-4 to collect the training set, though we will show in Table~\ref{tab:different_llms} how our model performs on ChatGPT and GPT-4 generated text. Hence, this paper uses GPT-3 for dataset generation, which serves as the benchmark for our measurements. Throughout this paper, the term ``GPT model'' specifically refers to GPT-3.

\begin{figure*}[t!]
\centering
\includegraphics[width=\textwidth]{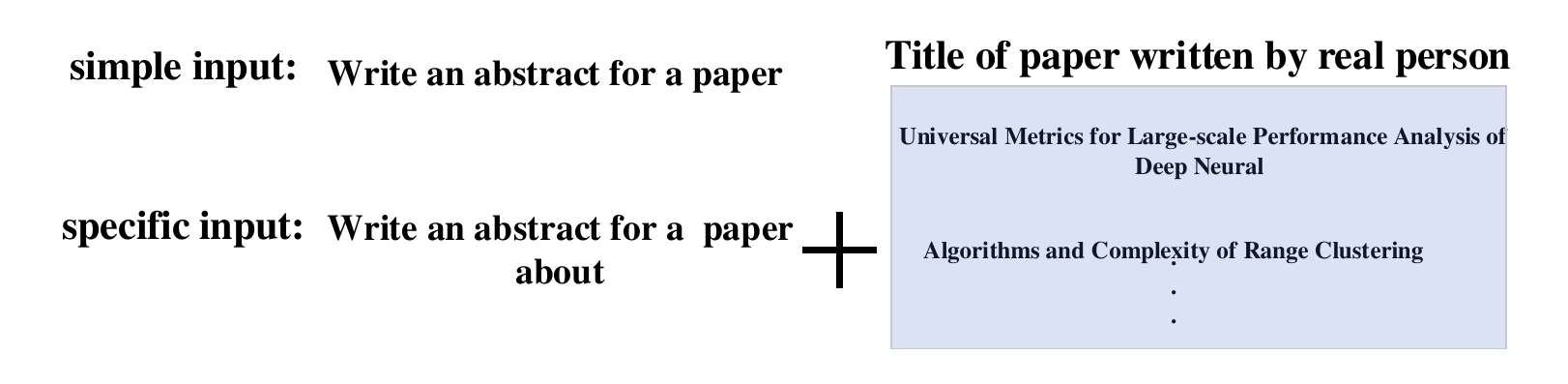}
\caption{Examples of a simple prompt and a specific prompt.}
\label{fig:simple_specific_prompts}
\end{figure*}

For the human-written part of the dataset, we collected 500 samples of paper abstracts about Artificial Intelligence written by actual authors from the arXiv dataset~\cite{clement2019use}, which is available on Kaggle\footnote{\url{https://www.kaggle.com/datasets/Cornell-University/arxiv}} and covers various fields. 
To create the AI-generated part of the dataset, we divided it into two subsets as depicted in Fig.~\ref{fig:simple_specific_prompts}. The ``Simple prompt'' subset consists of 500 GPT abstracts generated by GPT-3 using the prompt ``Write an abstract for a professional paper.'' The ``Specific prompt'' subset includes 500 GPT abstracts generated by GPT-3 using prompts beginning with "Write an abstract for a paper about" followed by the corresponding titles from the actual abstracts.

Among the state-of-the-art detectors available, such as ChatGPT detector and GPTZero, many lack associated published articles or datasets for reproduction. Moreover, a significant number of these detectors do not provide APIs, making it impossible to conduct batch-testing experiments. Consequently, we have chosen the RoBERTa base OpenAI Detector (OpenAI detector for short) on Hugging Face\footnote{\url{https://huggingface.co/roberta-base-openai-detector}}, a single-input binary classifier, as our target detector due to its availability and usability. At the same time, we also use the most widely used DetectGPT~\cite{mitchell2023detectgpt} as another target detector. The comparison of the performance between our model and DetectGPT, OpenAI detector can be seen in Table~\ref{tab:prompt_variants}.

The OpenAI detector demonstrates an exceptionally high accuracy of 98\% in detecting abstracts generated by \textbf{simple prompts} and 98\% in identifying human-written abstracts. However, for abstracts generated by \textbf{specific prompts}, the accuracy rate drops to 70. This substantial reduction in performance by simply adding a human-written sentence to the prompt clearly indicates the limited robustness of existing detectors. Notably, specific prompts are commonly used in academic cheating scenarios, where students tailor their assignments or reports to meet specific requirements provided by their professors, utilizing prompts similar to the specific prompts used in this study.

\section{Hypothesis for Prompt-Induced Lack of Robustness}
\label{sec::Hypothesis}
\begin{figure*}[t!]
\centering
\includegraphics[width=120mm]{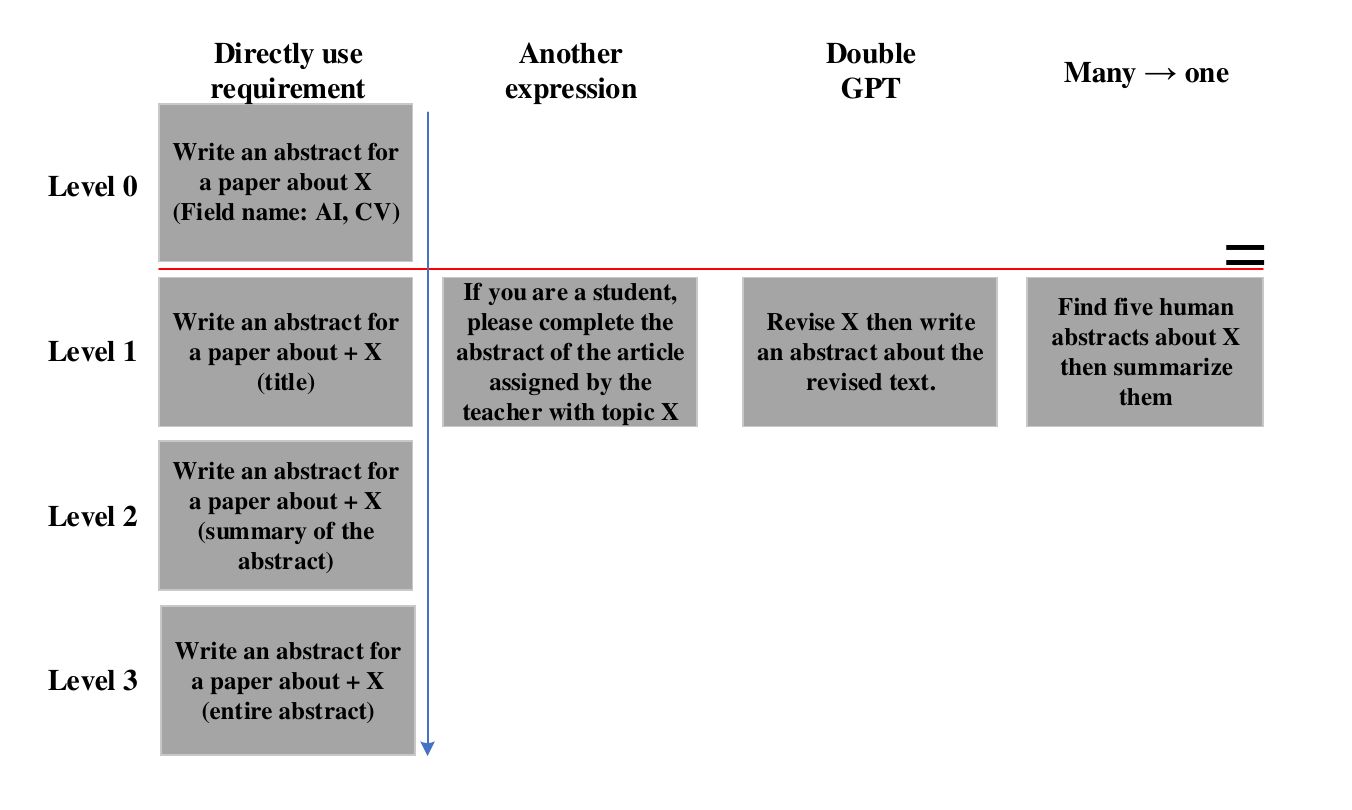}
\caption{The prompts can be categorized into different levels based on the length and complexity of the Content part and different Template variants based on how generation task is expressed in the Template part}
\label{fig:levels}
\end{figure*}

In our hypothesis, we first divide the prompt into two parts: the \textbf{Template} describing the task summary and the \textbf{Content} containing specific human information, which is denoted by \textbf{X}. The Content here means that in addition to the text describing the task user wants the LLM to do, the prompt also contains text with practical meaning, such as the title of an actual paper, written by real people.
\begin{itemize}
\item For the \textbf{Template} part, prompts for the same task can be divided into different Template variants depending on how task is expressed. Template variants include \textit{``Directly use requirement''} which is the specific prompt we designed before. \textit{``Another expression''} is where the student expresses the meaning of the requirement using different wording. The \textit{``Double GPT''} variant involves using the generation model (GPT) twice, where the student modifies Content X using GPT before generating the article. Lastly, the \textit{``Many $\rightarrow$ one''} variant simulates a common plagiarism method where the student collects five articles written by people about idea X and combines them into a new article. 
These Template variants allow us to evaluate the detector's performance in detecting different manipulative strategies employed by students.
Examples of each Template variant are shown in Fig.~\ref{fig:levels}
\item For the \textbf{Content} part, prompts can be divided into different levels depending on the length and complexity of the Content. The levels range from including only the field name to including the title, summary of the abstract, and the entire abstract, denoted as 0, 1, 2, and 3, respectively.

\end{itemize}

Our \textbf{hypothesis} can be explained as follows:
\begin{itemize}
\item Within the prompt, only component X, which is Content, influences the characteristics of the generated articles and contributes to the limited robustness observed in existing detectors.
\item When component X remains at a certain level, the generated articles exhibit similar characteristics regardless of the other parts of the prompt as shown by the red horizontal line in Fig.~\ref{fig:levels}.
\item As the complexity and level of detail in the X component increase, it becomes more challenging to detect the generated articles as shown by the blue arrow in Fig.~\ref{fig:levels}.
\end{itemize}

To verify this hypothesis, we designed the following experiments: First Fig.~\ref{fig:levels} contains a total of four variants and four levels, for a total of 16 squares. We use the prompt corresponding to each square to generate 100 machine-generated texts for a total of 16 sets of test data. Then, we use each set of data to test the accuracy of the OpenAI detector and fill in the corresponding position in Table~\ref{tab:human_contribution_result}. 

\begin{table*}
\centering
\caption{Accuracy of the original OpenAI detector (before fine-tuning) in different levels. X denotes the Content incorporated into the prompts.}
\label{tab:human_contribution_result}
\begin{tabular}{lrrrr} 
\hline
\textbf{X level} & \multicolumn{1}{c}{\begin{tabular}[c]{@{}c@{}}\textbf{Directly use }\\\textbf{requirement}\end{tabular}} & \multicolumn{1}{c}{\begin{tabular}[c]{@{}c@{}}\textbf{Another}\\\textbf{expression}\end{tabular}} & \multicolumn{1}{c}{\begin{tabular}[c]{@{}c@{}}\textbf{Double}\\\textbf{GPT}\end{tabular}} & \multicolumn{1}{c}{\textbf{Many $\rightarrow$ one}} \\ 
\hline
\textbf{level 0} (X = Field name) & 100\% & 100\% & 99\% & 86\% \\ 
\textbf{level 1} (X = Title) & 70\% & 74\% & 53\% & 72\% \\ 
\textbf{level 2} (X = Summary of abstract) & 34\% & 24\% & 20\% & 29\% \\ 
\textbf{level 3} (X = Entire abstract) & 11\% & 17\% & 7\% & 11\% \\
\hline
\end{tabular}
\end{table*}

The results in Table~\ref{tab:human_contribution_result} are consistent with our hypothesis. In summary, the X component of the prompt plays a crucial role in the characteristics of the generated articles and poses challenges for detection, particularly as it becomes more intricate and detailed. 
At the same time, the results in Table~\ref{tab:human_contribution_result} also show how the specific prompt affects the accuracy of the OpenAI detector. 
The accuracy remained acceptable at level 1 but deteriorated significantly at \textbf{level 2} and above. 



\section{Synthetic-Siamese}
\label{sec:Synthetic-Siamese}
Synthetic-Siamese consists of two key components. First, we analyze potential academic cheating scenarios and develop a \textbf{cheating model} specifically tailored to address these instances of cheating. Then we propose a \textbf{detection system} designed to identify instances of academic cheating based on our developed model.

\subsection{Student Cheating Model}
As depicted in Fig. \ref{fig:Model and System}, 
the model comprises two parties: the student side and the teacher side. The teacher assigns specific topic including Template and Content (defined in Sect.~\ref{sec::Hypothesis}) for an academic task, and then a potentially deceitful student utilizes these topic as prompt (as described by (2) in Fig.~\ref{fig:Model and System}) to generate an article using a generation model such as GPT-3. 
Meanwhile, the teacher also proactively employs the generation model to generate an article. 
Then, the teacher uses a model to compare the similarities between the student's article and their own generated article to determine whether the student has cheated.


This cheating model closely resembles real-life situations where students' assignments or examinations are typically centered around specific area and come with detailed requirements from teachers. To meet these requirements, students generally use the teacher's topic as input for generating their articles.
However, it is also possible that students do not directly use the input prompt same as the teacher's. 
For example, students may use their own topic different from teacher’s as the Content of prompts (which means that the teacher may not know the level of the Content of the prompt that students may use.) 
or modify the teacher's Template to use a different way of describing the generation tasks to build students' own prompts (as described by (1) in Fig.~\ref{fig:Model and System}). 
To verify that our model can deal with various real-life scenarios, we designed a \textbf{Level generalization test} and a \textbf{scenario generalization test} in Sect.~\ref{Sec:Experiment}

\subsection{Detection System}

\begin{figure*}[t!]
\centering
\includegraphics[width=120mm]{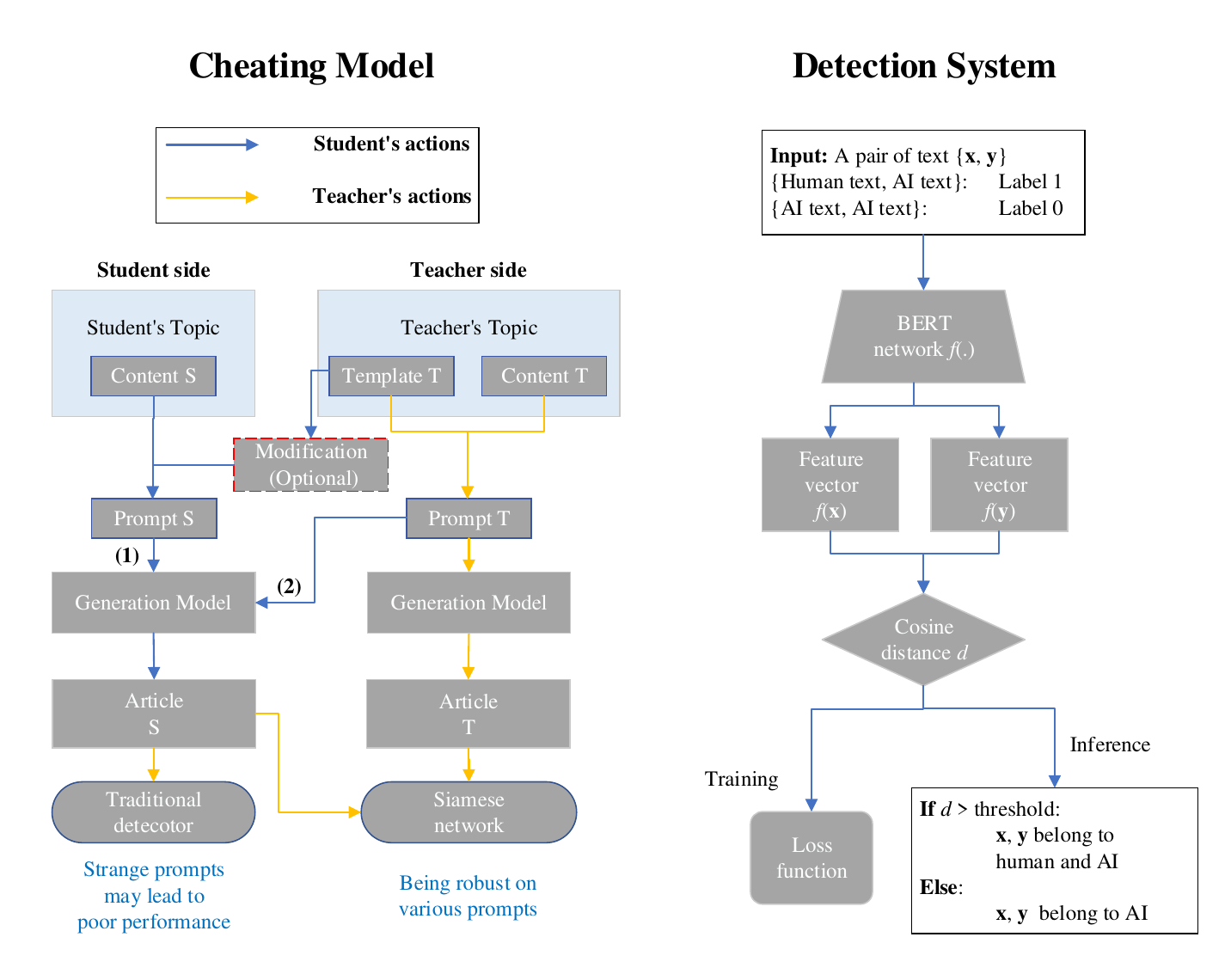}
\caption{The figure on the left is the Cheating Model, and the figure on the right is the Detection System. In the Cheating Model, both the student and teacher sides need to use Prompt containing Content and Template (defined in Sect.~\ref{sec::Hypothesis}) as input to the generation model and generate corresponding articles. We use the two suffixes S and T to distinguish the prompts students and teachers use and the articles they get.
}
\label{fig:Model and System}
\end{figure*}

As depicted in Fig.~\ref{fig:Model and System}, the network structure involves the input of two articles, \textbf{x} and \textbf{y}. Article \textbf{y} represents the teacher's AI-generated article, while \textbf{x} can either be a human-written article or an AI-generated one submitted by the student. 

Our detector employs a pre-trained BERT network as a feature extractor, denoted as $f(.)$, which is initialized with pre-trained weights. We fine-tune it using a supervised training approach. When labeling training data, if both \textbf{x} and \textbf{y} represent AI-generated articles, the label $l$ is assigned as 0. Conversely, if \textbf{x} corresponds to a human-written article and \textbf{y} represents an AI-generated article, the label $l$ is set as 1.

We use cosine distance $d(.,.)$ to measure the similarity between two feature vectors $\mathbf{f_x} = f(\mathbf{x})$ and $\mathbf{f_y} = f(\mathbf{y})$, as described in Eq.~\ref{eq:cosine}.

\begin{equation}
\label{eq:cosine}
 d(\mathbf{f_x}, \mathbf{f_y}) = 1 - \frac{\mathbf{f_x} \cdot \mathbf{f_y}}{\lVert\mathbf{f_x}\rVert_2 \lVert\mathbf{f_y}\rVert_2}
\end{equation}

The loss function utilized during training is described by Eq.~\ref{eq:loss}.
\begin{equation}
\label{eq:loss}
 \mathcal{L} = ld(\mathbf{f_x}, \mathbf{f_y})^2 + (1 - l)(2 - d(\mathbf{f_x}, \mathbf{f_y}))^2
\end{equation}

During the inference phase, our model calculates the cosine distance between the two input texts. A smaller distance indicates a higher similarity between \textbf{x} and \textbf{y}. As \textbf{y} represents AI-generated text, a smaller distance indicates that \textbf{x} is more likely to be generated by AI. Conversely, \textbf{x} is more likely to be written by a real person.

For the practicality of this detection system, it need two inputs: the article to be detected from students and the generated GPT article from teachers themselves. 
However, in most cases, Teacher's Topic are easily accessible to teachers themselves. At the same time, teachers can easily use generation models (such as ChatGPT) with low learning costs. Therefore, this detection system is practical in academic cheating.

\section{Experiment Results and Discussions}
We conducted the following experiments with an expanded dataset as outlined below to show that Synthetic-Siamese can achieve higher accuracy than existing detectors in various generalization scenarios.
\label{Sec:Experiment}
\subsection{Experimental Design}
The model settings for Synthetic-Siamese are as follows:
\begin{itemize}
\item The training set for Synthetic-Siamese consisted of 2,000 human-written abstracts from arXiv about Artificial Intelligence and 4,000 GPT-3 generated abstracts using \textbf{level 1} specific prompts. This dataset was also employed for fine-tuning the OpenAI detector. \textbf{Content} and \textbf{Template} mentioned in the following are described in Sect.~\ref{sec::Hypothesis}.

\item For \textbf{teacher's articles}. As described in Sect.~\ref{sec:Synthetic-Siamese}, Synthetic-Siamese requires two inputs, which are represented by (x, y) in Fig~\ref{fig:Model and System}: one is the article submitted by the student (the text to be detected), and the other is the article generated by the teacher. Therefore, when testing Synthetic-Siamese, in addition to the text to be detected, we also need to provide Synthetic-Siamese with machine-generated text as the teacher's article.
As a result, we generated 100 GPT-3 abstracts using the prompts of only \textit{``Directly use requirement''} Template variant in level 1 as the teacher's article. We use level 1 because, in real-life scenarios, no matter what the teacher’s topic is and whether it is specific or detailed, it is enough to serve as the Content of level 1 prompts. Therefore, in the following tests, we use this data set as the teacher's article.
\end{itemize}
For student articles (the text to be detected). We designed the following three generalization test sets:
\begin{itemize}
\item For the \textbf{level generalization test set}, we selected 100 human-written abstracts and generated 100 GPT abstracts using prompts in level 1, 2 and 3 each with the \textit{``Directly use requirement''} Template variant that mimics different manipulative behaviors students may employ.
This test set helps us verify that Synthetic-Siamese can maintain high robustness in all levels while the teacher only use the text generated by prompts in level 1 as the teacher's article and does not need to know the prompt level of the student's article.

\item For the \textbf{scenario generalization test set}, 
we selected 100 human-written abstracts and generated 100 GPT abstracts using prompts with four different Template variants each in level 1 as the student's article which need to be detected to simulate how a student might evade detection by describing the task in a different way. 
In addition to the prompt of teacher article and the student article using different Template variants, 
We also aim to make the prompt of the two use different Content. 
Therefore, to simulate a student evading detection by using different human information from the teacher's topic. This test set helps us verify that Synthetic-Siamese can maintain high robustness when student attempts to evade detection.

\item For the \textbf{model generalization test set}, we chose 50 human-written abstracts and generated 50 abstracts for each generation model comprising OpenAI's GPT-3, Perplexity's customized GPT-3.5\footnote{\url{https://www.perplexity.ai/}}, the Falcon-7B\footnote{\url{https://falconllm.tii.ae/}}, ChatGPT, and GPT-4. All abstracts were generated using level 1 and level 2 prompts. This test set enables us to assess the Synthetic-Siamese's ability to generalize across different generation models, providing insights into their performance and adaptability in diverse AI-generated text scenarios.
\end{itemize}


\subsection{Level Generalizability}
\label{}
In level generalization test set, we simulated that teachers always use prompts in level 1, while students use level 1, 2 and 3.
As presented in Table~\ref{tab:prompt_variants}, the OpenAI detector exhibited a significant drop in performance of \textbf{level 3} prompts, even after fine-tuning, with a maximum accuracy of only 66.5\%. 
Although DetectGPT is a widely used detector now, its maximum accuracy in level 3 is only 50.5\%. This is because DetectGPT scores the text to be detected and determines whether the text was written by a person or GPT based on whether the score is greater than 50\% (the boundary between humans and machines). As a result, in certain situations (such as the text generated by the specific prompt mentioned in this paper), DetectGPT tends to produce ambiguous results, i.e., the score is in the middle. 

In contrast, Synthetic-Siamese demonstrated much greater generalizability, achieving the accuracy of 95\% in all three levels. This level generalizability demonstrates that in academic cheating scenarios, Synthetic-Siamese can effectively detect the students’ usage of GPT, regardless of the level of Content that students use.

\begin{table}
\centering
\caption{Accuracy of the detectors on the Level generalization test set with level 1, 2 and 3 prompts.}
\label{tab:prompt_variants}
\begin{tabular}{lrrrrr} 
\hline
\begin{tabular}[c]{@{}l@{}}\textbf{Level of}\\\textbf{the prompt}\end{tabular} & \begin{tabular}[c]{@{}r@{}}\textbf{OpenAI}\\\textbf{detector}\\\textbf{(original)}\end{tabular} & \begin{tabular}[c]{@{}r@{}}\textbf{OpenAI}\\\textbf{detector}\\\textbf{(fine-tuned)}\end{tabular} & \begin{tabular}[c]{@{}r@{}}\textbf{DetectGPT}\\\textbf{(original)}\end{tabular} & \begin{tabular}[c]{@{}r@{}}\textbf{Synthetic-}\\\textbf{Siamese}\end{tabular} & \begin{tabular}[c]{@{}r@{}}\textbf{Single-input}\\\textbf{Model}\end{tabular} \\ 
\hline
Level 1 & 85.0\% & 98.5\% & 100.0\% & 95.0\% & 99.0\% \\ 
\hline
Level 2 & 66.5\% & 89.0\% & 50.5\% & 95.0\% & 85.5\%\\ 
\hline
Level 3 & 54.5\% & 66.5\% & 50.5\% & 95.0\% & 58.5\%\\ 
\hline
\end{tabular}
\end{table}

At the same time, to prove that our proposed Synthetic-Siamese has better level generalizability than traditional single-input network detectors. Using the same training set, we trained a single-input traditional classification model (with the same Bert as Synthetic-Siamese following a three-layer classifier). As presented in Table~\ref{tab:prompt_variants}, the performance of the single-input model is worse than that of Synthetic-Siamese. As a result, the Synthetic-Siamese model structure is the root cause of better level generalizability.


\subsection{Scenario Generalizability}
In the scenario generalization test set, we simulated possible methods that students may use to avoid detection when using GPT. We separately tested the accuracy of Synthetic-Siamese when the student used the same Content as the teacher's prompt but a different Template than the teacher’s, and when the student used both different Content and Template than the teacher. 
As presented in Table~\ref{tab:scenario_variants}, regardless of whether the Content or the Template of prompts used by students or teachers are consistent, Synthetic-Siamese maintains high accuracy. 

Template and Content in the Prompt will affect the generated text under human observation. But for models, they don't affect the distinction between human-written and machine-generated.
This scenario generalizability demonstrates that Synthetic-Siamese can still perform well even if the student uses different prompts than the teacher to evade detection.

\begin{table*}
\centering
\caption{Accuracy of Synthetic-Siamese on the scenario-generalization test set. The Template variant of prompts used by teacher is \textit{``Directly use requirement''} while the student use four different Template variants. The upper line represents the result when Content of teacher's prompts and student's prompts are same, while the lower line represents the result when they are different.}
\label{tab:scenario_variants}
\begin{tabular}{c|rrrr} 
\hline
\multirow{2}{*}{\begin{tabular}[c]{@{}c@{}}\textbf{Content of}\\ \textbf{Teacher's and Student's Prompt}\end{tabular}} & \multicolumn{4}{c}{\textbf{Template variant of Student Prompt}}\\ \cline{2-5} & \multicolumn{1}{c}{\begin{tabular}[c]{@{}c@{}}\textbf{Directly use }\\\textbf{requirement}\end{tabular}} & \multicolumn{1}{c}{\begin{tabular}[c]{@{}c@{}}\textbf{Another}\\\textbf{expression}\end{tabular}} & \multicolumn{1}{c}{\begin{tabular}[c]{@{}c@{}}\textbf{Double}\\\textbf{GPT}\end{tabular}} & \multicolumn{1}{c}{\textbf{Many $\rightarrow$ one}} \\
\hline
Same & 95.0\% & 95.0\% & 95.0\% & 95.0\% \\ 
Different & 95.0\% & 95.0\% & 95.0\% & 95.0\% \\ 
\hline
\end{tabular}
\end{table*}


\subsection{Model Generalizability}
Although GPT has become mainstream, students may utilize several other LLM-based text-generation models to avoid detection. To assess Synthetic-Siamese's effectiveness, we conducted tests using the model generalization test set. It is important to note that teacher's articles were only generated by GPT-3.

The results of the tests are presented in Table~\ref{tab:different_llms}. The DetectGPT and the original OpenAI detector struggled to perform effectively in most cases, while the fine-tuned OpenAI detector achieved the better accuracy except when dealing with text generated by Falcon-7B using \textbf{level-2} prompts. Synthetic-Siamese performed highly on the GPT-3 and customized GPT-3.5 but showed limited generalizability when faced with Falcon's generated text. 
We hypothesized that when training our Siamese-based detector with the proposed cheating model, the detector learned to identify authorship information. It distinguished GPT as one author and humans as another. When a new ``author'' (Falcon-7B) emerged, the detector struggled to assign its text to either human or GPT. 


For ChatGPT and GPT-4, the performance of Synthetic-Siamese is better than the other three detectors. Traditional detectors such as OpenAI detector are more suitable for \textbf{closed set detection}, so if we use level 1 prompts to fine-tune the OpenAI detector and then use level 1 to test, the performance of the OpenAI detector can significantly improve. However, for level 2, the performance of the fine-tuned OpenAI detector drops significantly. In contrast, Synthetic-Siamese has excellent \textbf{cross-level open set detection} capabilities. Regardless of the level, Synthetic-Siamese maintains high accuracy. This model generalizability demonstrates that Synthetic-Siamese can effectively detect the students’ usage of generation models, no matter which model students use.


\begin{table}
\centering
\caption{Accuracy of the detectors on the text generated by different LLMs. OpenAI detector, a binary classifier, only needs one input. Besides the query text, our detector requires the corresponding generated text (from the teacher) as an anchor. Within each cell, the upper number represents the result on \textbf{level-1} prompts, while the lower number represents the result on \textbf{level-2} prompts.}
\label{tab:different_llms}
\begin{tabular}{l|rrrrr} 
\hline
\multirow{2}{*}{\begin{tabular}[c]{@{}l@{}}\textbf{Detector}\end{tabular}}  & \multicolumn{5}{c}{\textbf{Source of input text}} \\ 
\cline{2-6}
 & \textbf{GPT-3} & \textbf{Falcon-7B} & \textbf{Perplexity} & \textbf{ChatGPT} & \textbf{GPT-4} \\
\hline
Synthetic-Siamese & \begin{tabular}[c]{@{}r@{}}92.0\%\\92.0\%\end{tabular} & \begin{tabular}[c]{@{}r@{}}88.0\%\\48.0\%\end{tabular} & \begin{tabular}[c]{@{}r@{}}100.0\%\\100.0\%\end{tabular} & \begin{tabular}[c]{@{}r@{}}100.0\%\\79.0\%\end{tabular} & \begin{tabular}[c]{@{}r@{}}100.0\%\\100.0\%\end{tabular} \\
\hdashline
OpenAI detector (original) & \begin{tabular}[c]{@{}r@{}}85.0\%\\66.5\%\end{tabular} & \begin{tabular}[c]{@{}r@{}}71.5\%\\55.5\%\end{tabular} & \begin{tabular}[c]{@{}r@{}}80.0\%\\74.0\%\end{tabular} & \begin{tabular}[c]{@{}r@{}}60.5\%\\53.5\%\end{tabular} & \begin{tabular}[c]{@{}r@{}}63.0\%\\54.0\%\end{tabular} \\
\hdashline
OpenAI detector (fine-tuned)  & \begin{tabular}[c]{@{}r@{}}98.5\%\\89.0\%\end{tabular} & \begin{tabular}[c]{@{}r@{}}90.5\%\\54.5\%\end{tabular} & \begin{tabular}[c]{@{}r@{}}98.5\%\\98.5\%\end{tabular} & \begin{tabular}[c]{@{}r@{}}98.5\%\\56.0\%\end{tabular} & \begin{tabular}[c]{@{}r@{}}97.0\%\\83.5\%\end{tabular}\\
\hdashline
DetectGPT & \begin{tabular}[c]{@{}r@{}}55.5\%\\52.5\%\end{tabular} & \begin{tabular}[c]{@{}r@{}}50.0\%\\50.0\%\end{tabular} & \begin{tabular}[c]{@{}r@{}}64.0\%\\51.0\%\end{tabular} & \begin{tabular}[c]{@{}r@{}}51.5\%\\50.0\%\end{tabular} & \begin{tabular}[c]{@{}r@{}}52.0\%\\52.5\%\end{tabular} \\
\hline
\end{tabular}
\end{table}

\section{Conclusion}
This study addresses the issue of academic cheating facilitated by LLMs, which are widely utilized in contemporary contexts. By examining the RoBERTa Base OpenAI Detector as a case study, we identified potential limitations in the robustness of existing detection methods. Additionally, we conducted an in-depth analysis and presented a hypothesis highlighting the role of Content (X factor) in prompts contributing to the detector's lack of robustness. We then formulated a cheating scenario in academic writing and proposed Synthetic-Siamese, a detection approach which determine whether a student cheated based on the similarity of the teacher's and student's texts. Our experimental results conclusively demonstrated that Synthetic-Siamese exhibits much greater prompt generalization capabilities compared to that of the OpenAI detector. 

\section*{Acknowledgments}
This work was partially supported by JSPS KAKENHI Grant JP21H04907, and by JST CREST Grants JPMJCR18A6 and JPMJCR20D3, Japan.

\bibliographystyle{plain} 
\bibliography{author_v1}

\begin{thebibliography}{10}

\bibitem{anil2023palm}
Rohan Anil, Andrew~M Dai, Orhan Firat, Melvin Johnson, Dmitry Lepikhin, Alexandre Passos, Siamak Shakeri, Emanuel Taropa, Paige Bailey, Zhifeng Chen, et~al.
\newblock {PaLM} 2 technical report.
\newblock {\em arXiv preprint arXiv:2305.10403}, 2023.

\bibitem{clement2019use}
Colin~B Clement, Matthew Bierbaum, Kevin~P O'Keeffe, and Alexander~A Alemi.
\newblock On the use of arxiv as a dataset.
\newblock {\em arXiv preprint arXiv:1905.00075}, 2019.

\bibitem{crothers2023machine}
Evan Crothers, Nathalie Japkowicz, and Herna Viktor.
\newblock Machine generated text: A comprehensive survey of threat models and detection methods, 2023.

\bibitem{greshake2023youve}
Kai Greshake, Sahar Abdelnabi, Shailesh Mishra, Christoph Endres, Thorsten Holz, and Mario Fritz.
\newblock Not what you've signed up for: Compromising real-world llm-integrated applications with indirect prompt injection, 2023.

\bibitem{jawahar2020automatic}
Ganesh Jawahar, Muhammad Abdul-Mageed, and Laks V.~S. Lakshmanan.
\newblock Automatic detection of machine generated text: A critical survey, 2020.

\bibitem{liyanage2022benchmark}
Vijini Liyanage, Davide Buscaldi, and Adeline Nazarenko.
\newblock A benchmark corpus for the detection of automatically generated text in academic publications, 2022.

\bibitem{mitchell2023detectgpt}
Eric Mitchell, Yoonho Lee, Alexander Khazatsky, Christopher~D. Manning, and Chelsea Finn.
\newblock Detectgpt: Zero-shot machine-generated text detection using probability curvature, 2023.

\bibitem{openaigpt4}
OpenAI.
\newblock {GPT-4} technical report.
\newblock {\em arXiv preprint arXiv:2303.08774}, 2023.

\bibitem{pu2022deepfake}
Jiameng Pu, Zain Sarwar, Sifat~Muhammad Abdullah, Abdullah Rehman, Yoonjin Kim, Parantapa Bhattacharya, Mobin Javed, and Bimal Viswanath.
\newblock Deepfake text detection: Limitations and opportunities, 2022.

\bibitem{fine-tunning}
Juan Rodriguez, Todd Hay, David Gros, Zain Shamsi, and Ravi Srinivasan.
\newblock Cross-domain detection of {GPT}-2-generated technical text.
\newblock In {\em Proceedings of the 2022 Conference of the North American Chapter of the Association for Computational Linguistics: Human Language Technologies}, pages 1213--1233, Seattle, United States, July 2022. Association for Computational Linguistics.

\bibitem{disinformation}
Harald Stiff and Fredrik Johansson.
\newblock Detecting computer-generated disinformation.
\newblock {\em International Journal of Data Science and Analytics}, 13, 05 2022.

\bibitem{yuan2021bartscore}
Weizhe Yuan, Graham Neubig, and Pengfei Liu.
\newblock {BARTScore}: Evaluating generated text as text generation, 2021.

\bibitem{zhao2023survey}
Wayne~Xin Zhao, Kun Zhou, Junyi Li, Tianyi Tang, Xiaolei Wang, Yupeng Hou, Yingqian Min, Beichen Zhang, Junjie Zhang, Zican Dong, Yifan Du, Chen Yang, Yushuo Chen, Zhipeng Chen, Jinhao Jiang, Ruiyang Ren, Yifan Li, Xinyu Tang, Zikang Liu, Peiyu Liu, Jian-Yun Nie, and Ji-Rong Wen.
\newblock A survey of large language models, 2023.

\end{thebibliography}

\end{document}